\pdfoutput=1
\documentclass[sigconf]{acmart}
\AtBeginDocument{%
  }

\setcopyright{none}
\acmConference[KDD 2026]{ACM SIGKDD Conference on Knowledge Discovery and Data Mining}{August 09--13, 2026}{Jeju, Korea}
\usepackage{fancyhdr}
\fancypagestyle{plain}{%
  \fancyhf{}%
  \fancyhead[LE,RO]{\thepage}%
}
\makeatletter
\def\ACM@acmISBN{}
\acmDOI{10.1145/3770855.3819018}
\makeatother
\usepackage{subcaption}
\usepackage{microtype}
\usepackage[table]{xcolor}
\usepackage{colortbl}
\definecolor{headercolor}{RGB}{230, 240, 255} 
\definecolor{sotacolor}{RGB}{245, 245, 250} 
\definecolor{teachercolor}{RGB}{255, 245, 240} 
\definecolor{guardcolor}{RGB}{240, 255, 240} 

\begin{document}

\title{When to Trust, How to Distill: Multi-Foundation Model Guidance for Lightweight, Robust Scientific Time Series Forecasting}

\author{Rupasree Dey}
\email{rupasree.dey@colostate.edu}
\orcid{0009-0001-4674-5089}
\affiliation{%
\department{Department of Computer Science}
  \institution{Colorado State University}
  \city{Fort Collins}
  \state{Colorado}
  \country{USA}
}

\author{Abdul Matin}
\email{abdul.matin@colostate.edu}
\orcid{0000-0002-2680-3544}
\affiliation{%
\department{Department of Computer Science}
  \institution{Colorado State University}
  \city{Fort Collins}
  \state{Colorado}
  \country{USA}
}

\author{Nathan Orwick}
\email{nathan.orwick@colostate.edu}
\orcid{0009-0002-7154-8164}
\affiliation{%
\department{Department of Computer Science}
  \institution{Colorado State University}
  \city{Fort Collins}
  \state{Colorado}
  \country{USA}
}

\author{Yao Zhang}
\email{yao.zhang@colostate.edu}
\orcid{0000-0002-2669-1378}
\affiliation{%
\department{Department of Soil and Crop Sciences}
  \institution{Colorado State University}
  \city{Fort Collins}
  \state{Colorado}
  \country{USA}
}

\author{Shrideep Pallickara}
\email{shrideep.pallickara@colostate.edu}
\orcid{0009-0004-4839-2493}
\affiliation{%
\department{Department of Computer Science}
  \institution{Colorado State University}
  \city{Fort Collins}
  \state{Colorado}
  \country{USA}
}

\author{Sangmi Lee Pallickara}
\email{sangmi.pallickara@colostate.edu}
\orcid{0000-0001-7012-5528}
\affiliation{%
\department{Department of Computer Science}
  \institution{Colorado State University}
  \city{Fort Collins}
  \state{Colorado}
  \country{USA}
}

\renewcommand{\shortauthors}{Dey et al.}

\begin{abstract}
The deployment of Time-Series Foundation Models (TSFMs) in
physical sciences is hindered by a critical trade-off: while these
models encode rich, universal temporal dynamics, they suffer from
severe distributional misalignment when applied zero-shot to
specific scientific domains, and their computational cost prohibits
deployment in edge-computing sensor networks. We address a
fundamental challenge: How can we extract latent structural
knowledge from misaligned foundation models (FM) to train lightweight,
specialized forecasters? We propose \textbf{G}ated \textbf{U}ncertainty-\textbf{A}ware \textbf{R}outing for \textbf{D}istillation (\textbf{\textsc{Guard}}), a novel framework that reframes multi-teacher distillation as an instance-wise decision process with two adaptive
mechanisms: (1) a Contextual Router that dynamically selects the
most relevant teacher based on local input statistics, exploiting
complementarity across diverse foundation models; and (2) an
Uncertainty-Gated Temperature mechanism that acts as a
``circuit-breaker,'' automatically attenuating distillation strength
when teacher confidence diverges from domain reality. We evaluate our proposed lightweight framework on four climate-critical domains: meteorology, ecosystem carbon flux, soil moisture, and energy grids. Our method significantly reduces RMSE relative to a fixed-weight multi-teacher distillation baseline, successfully distilling knowledge from pretrained FMs (teachers) even when they exhibit suboptimal zero-shot accuracy due to distribution shift between the original and target data domains. We demonstrate that these domain-misaligned teachers can still serve as critical correctives, outperforming the globally superior FMs on 28.5\% of the hardest instances. Ultimately, this enables high-precision scientific forecasting suitable for resource-constrained edge deployment.
Code is available at \url{https://github.com/RupasreeDey/GUARD-KDD2026}.
\end{abstract}




\maketitle
\thispagestyle{plain} 

\section{Introduction}
Accurate and resilient time-series forecasting is critical to understand natural phenomena and to monitor the performance of long-lasting infrastructures such as meteorological systems, ecosystem dynamics, and energy grid operations. However, capturing complex temporal trends and patterns using conventional modeling strategies presents significant challenges due to inherent label scarcity, high-stakes data acquisition costs, and the complexity of ancillary factors influencing system behavior ~\citep{kim2025comprehensive, yang2025not, du2021adarnn}. 

Recent machine learning advances in time-series foundation models (TSFMs) have demonstrated strong potential with their generalizability across diverse domains ~\citep{das2024decoder, ansari2024chronos, rasul2023lag, woo2024unified}. TSFMs are large-scale models trained on voluminous and heterogeneous temporal corpora. While these models capture rich, universal temporal dynamics (e.g., seasonality and trends), their direct deployment in scientific workflows via zero-shot inference is often infeasible ~\citep{meyer2025time}. In particular, applying TSFMs to new or specialized scientific phenomena presents significant challenges in maintaining acceptable performance without task or domain specific model refinement ~\citep{trivsovic2025rapid}.


Fine-tuning billion-parameter models is computationally prohibitive for most scientific labs \citep{trivsovic2025rapid}. Several approaches have been proposed to develop efficient, robust, and domain-adaptive methods that transfer the latent temporal reasoning of FMs into lightweight, task-specific forecasters. Knowledge distillation ~\citep{hinton2015distilling}, which extracts learned representations and insights from a high-capacity teacher model into a compact student model.  However, conventional distillation strategies typically treat the teacher model’s outputs as target labels, an assumption that becomes problematic when the student model operates on data whose scope, distribution, or modality is misaligned with that used to train the teacher. Such data misalignment is common in environmental monitoring, where new observations are continuously assimilated and system dynamics are strongly influenced by ancillary conditions. These factors widen the gap between the knowledge encoded in the teacher and what the student can effectively learn. As a result, the student models often struggle to capture critical temporal patterns, including extreme events and prolonged calm periods ~\citep{himtm_hierarchical, zhang2025shiftkd}.

To address these challenges, we introduce \textbf{\textsc{Guard}} (\textbf{G}ated \textbf{U}n\-certainty-\textbf{A}ware \textbf{R}outing for \textbf{D}istillation), a framework designed to distill insights from multiple TSFM teacher models by leveraging their complementary strengths.
While Ensembling requires running all giant teachers at inference time (expensive) our method compiles them into one tiny student (lightweight). Our novel knowledge distillation framework provides instance-wise dynamic decision-making throughout the distillation process. Rather than blindly mimicking a single teacher, our framework adaptively orchestrates a multi-faceted learning process based on the estimated reliability of each teacher model for a given training sample.

\textbf{\textsc{Guard}} employs two adaptive mechanisms. First, a contextual router 
dynamically weights teachers based on local input statistics, enabling 
sample-specific composition of multiple TSFMs. Second, an adaptive temperature 
network rescales distillation strength based on teacher uncertainty ~\citep{guo2024leveraging}, acting as 
a 'circuit breaker' that filters unreliable guidance. This attenuates harmful gradients from misaligned teachers while preserving robust temporal priors such as seasonality and trends.

We evaluate \textbf{\textsc{Guard}} using two widely adopted TSFMs: TimesFM ~\citep{das2024decoder}, a large-scale time-series foundation model trained on heterogeneous real-world temporal data, and Chronos ~\citep{ansari2024chronos}, a probabilistic transformer-based forecasting model pretrained on massive synthetic and real time-series corpora. These two models demonstrate complementary performance across different data distributions (see Section 3). Experiments are conducted on forecasting tasks spanning four scientific domains: meteorology (Weather), ecosystem carbon fluxes (Flux), and electrical transformer load forecasting (ETTm1 and ETTh1). The key contributions of this work are as follows:




\noindent
\textbf{1. Dynamic Orchestration:} \textsc{Guard} dynamically orchestrates multi-teacher knowledge distillation through an instance-wise adaptive weighting mechanism, enabling the student to adapt to the heterogeneous dynamics of scientific time series.

\noindent
\textbf{2. Uncertainty-Aware Gating:} \textsc{Guard} adaptively adjusts distillation strength per training sample based on estimated teacher uncertainty, extracting useful structural priors while filtering noise from distributional misalignment.

\noindent
\textbf{3. Efficient Edge Deployment:} \textsc{Guard} produces a lightweight student ($\sim$0.3M params) that achieves state-of-the-art accuracy on scientific benchmarks by synergistically leveraging complementary—yet imperfect—foundation model signals.

\section{Related Work}
\label{sec:related_work}

\textbf{Time Series Forecasting and Distillation}
Knowledge Distillation (KD)~\citep{hinton2015distilling} has been proven effective 
for model compression in computer vision~\citep{gou2021knowledge} and 
NLP~\citep{sanh2019distilbert}. Recent work has extended KD to time-series forecasting, 
typically employing single-teacher frameworks with lightweight student 
architectures~\citep{floratos2022online, ni2025timedistill, 
liu2025efficient, liu2024taming}. However, these 
methods assume consistent teacher quality across all samples—an assumption 
that breaks down in scientific domains where distribution shift is endemic 
and sensor noise is stochastic. Moreover, existing time-series KD approaches 
rarely leverage the complementary strengths of multiple foundation models, 
instead relying on a single teacher that may be misaligned with the target 
domain.

\textbf{Multi-Teacher and Adaptive Learning}
Multi-teacher KD aggregates diverse teacher insights to produce more robust 
student models~\citep{furlanello2018born, wu-etal-2021-one}. Existing approaches employ various teacher selection 
strategies, including reinforcement learning~\citep{yang2025multi}, 
attention-based weighting~\citep{du2020agree}, 
and meta-learning~\citep{zhang2023adaptive}. Recent work has 
introduced temperature-based modulation~\citep{lin2024atmkd, 
long2024mkdat} and uncertainty-gating mechanisms~\citep{zhang2022confidence, 
wang2025adahet} to handle heterogeneous teacher quality. However, these methods lack mechanisms to detect 
severe distribution misalignment, overlook regime-dependent complementarity 
where weak teachers excel in specific regimes, and ignore temporal heterogeneity 
across prediction horizons.

\section{Preliminary Analysis}
\label{sec:preliminary}

Traditional knowledge distillation assumes teacher models provide reliable guidance across all inputs, but this breaks down when foundation models (FMs) are applied to scientific domains outside their pretraining distribution. To investigate, we analyze two widely-adopted time-series FMs, TimesFM and Chronos—on scientific weather data. Although both are pretrained on large-scale corpora, their architectural differences produce distinct, complementary failure modes.

\begin{table}[t]
\centering
\caption{Global complementarity statistics on the Weather validation set summarizing overall performance and conditional minority-teacher gains. Because TimesFM is globally dominant, conditional win-rate statistics focus on Chronos to quantify specialist behavior rather than symmetric win counts.}
\label{tab:preliminary-complementarity}
\small
\begin{tabular}{lp{1.2cm}p{3.8cm}}
\toprule
\textbf{Metric} & \textbf{Value} & \textbf{Interpretation} \\
\midrule
TimesFM mean RMSE & 0.86 & Globally dominant teacher \\
Chronos mean RMSE & 1.52 & Weaker overall performance \\
\midrule
Chronos win rate & 28.5\% & Conditional superiority in subset of windows \\
Error correlation ($\rho$) & 0.52 & Distinct failure modes \\
Hard sample win rate & 22.9\% & Gains on TimesFM's difficult cases \\
\bottomrule
\end{tabular}
\end{table}

We compared both teachers on weather observation forecasting tasks. As shown in Table~\ref{tab:preliminary-complementarity}, TimesFM achieves substantially lower overall error (RMSE $0.86$ vs.\ $1.52$), establishing it as the globally dominant teacher. Nevertheless, Chronos produces lower error in $28.5\%$ of validation windows, demonstrating meaningful conditional complementarity. The relatively low error correlation ($\rho = 0.52$) further indicates that the two teachers exhibit non-redundant failure modes. Because TimesFM dominates globally, we report conditional win statistics for Chronos to characterize minority-teacher specialization rather than symmetric win counts.

Volatility stratification reveals that Chronos's win rate rises to $35.8\%$ in calm regimes (std\,$<$\,0.22) but falls to $18.1\%$ in volatile ones (std\,$>$\,0.34), reflecting architectural differences between its quantile-binning formulation and TimesFM's regression approach (full analysis in Appendix~\ref{app:preliminary}). Signal magnitude and local volatility account for most of the discriminative power, confirming routing can be learned from lightweight observable statistics. These findings motivate adaptive, instance-wise distillation: a learned router approximates oracle routing from test-time features (volatility, magnitude, trend), while uncertainty-aware hedging provides robustness under domain shift.

\section{Methodology}
\label{sec:methodology}

\textsc{Guard} operationalizes the complementarity observed in Section~\ref{sec:preliminary} 
via two adaptive mechanisms: a \textit{Contextual Router} for regime-dependent teacher 
weighting, and an \textit{Uncertainty-Gated Temperature Network} that suppresses 
distillation when teachers are misaligned.

\subsection{Training Pipeline Overview}

Our framework follows a two-phase pipeline to distill complementary knowledge from multiple foundation model teachers into a lightweight student while avoiding repeated teacher inference.

\textbf{Phase 1: Teacher Inference and Caching.}
All pretrained teachers are executed in zero-shot mode over training, validation, and test context windows. For each window–horizon instance, we cache teacher predictions and uncertainty estimates. Using ground-truth targets during training, we compute pseudo-oracle routing weights based on relative teacher errors; these are used only as supervision for router training and are not used during inference.

\textbf{Phase 2: Joint Adaptive Distillation.}
The student, Contextual Router, and Adaptive Temperature Network are trained jointly using cached teacher outputs. Regime features and teacher uncertainties guide routing weights and adaptive temperatures, and weighted teacher predictions form unified distillation targets. Training optimizes a combination of supervised forecasting loss, uncertainty-aware distillation loss, and entropy regularization. At inference time, the lightweight student and routing components are used.

The following subsections detail each component of this pipeline, including regime feature construction, adaptive routing, uncertainty-aware distillation gating, and the joint optimization objective.

\subsection{Problem Formulation}

We consider multi-horizon forecasting on a univariate target series $\{y_t\}_{t=1}^{T}$ with multivariate covariates $\{\mathbf{x}_t\}_{t=1}^{T}$ where $\mathbf{x}_t \in \mathbb{R}^{d}$. Given a context window of length $L$, the student model $f_\theta$ predicts future values across multiple horizons as $\hat{y}{t+1:t+h_k} = f\theta(\mathbf{x}{t-L+1:t}, y{t-L+1:t})$ for all $h_k \in \mathcal{H}$.

Unlike standard supervised learning where all training labels are equally trustworthy, our setting requires meta-reasoning about teacher reliability. The student must simultaneously learn to forecast and to recognize when each teacher's guidance is valuable versus misleading.

\begin{figure*}
    \centering
    \includegraphics[width=0.95\linewidth]{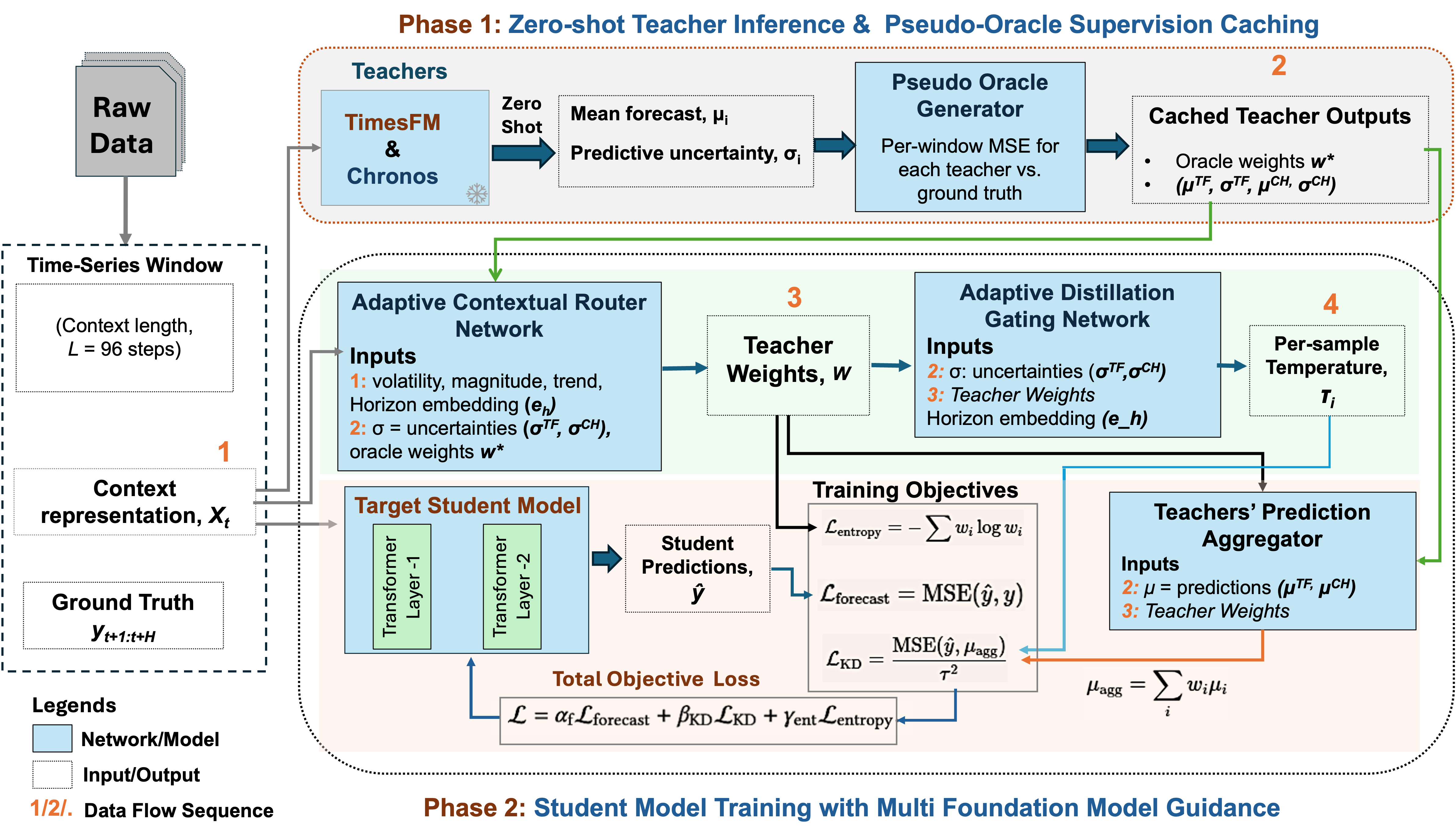}
    \caption{Overview of the Proposed Framework (\textsc{Guard}). \textbf{Phase-1:} Foundation models (Teachers) generate forecasts and uncertainty estimates, which are used to compute Oracle weights and cached. \textbf{Phase-2:} The Student model is trained via two adaptive paths: (1) a \textbf{Contextual Router Network} that predicts mixing weights $w$ based on local regime features ($s$), and (2) a \textbf{Temperature Network} that calibrates distillation strength $\tau$ based on teacher uncertainty. The final training objective combines forecast accuracy with uncertainty-aware distillation.}
    \Description{Methodology}
    \label{fig:meth}
\end{figure*}

\subsection{Construction of Local Regime Features}
The regime stratification observed in Sec.~\ref{sec:preliminary} highlights the strong correlation between temporal signal characteristics and model performance.To facilitate teacher models in learing local regimes, we characterize local signals by extracting three statistics over the 12 trailing timesteps: (1) rolling standard deviation (variability), (2) last observed value (signal magnitude, accounting for 68\% of selection importance), and (3) local linear trend (directional context). These domain-agnostic features are z-normalized using training statistics. Data is split chronologically into sliding windows of length $L$ to construct context $\mathbf{X}^{(i)} \in \mathbb{R}^{L \times d}$ and future targets $\mathbf{y}^{(i)}_{(h)} = [y_{i+1}, \dots, y_{i+h}]$.

\subsection{Zero-shot Teacher Inference \& Pseudo-Oracle Supervision Caching}

In this study, we employ TimesFM (continuous regression) and Chronos (quantile-based) as complementary teachers, because their architectural differences produce the clear error decorrelation ($\rho$= 0.52), making this combination well suited to demonstrate the effectiveness of our selective distillation approach. Each teacher generates predictions and uncertainties for every window: $(\boldsymbol{\mu}^{\text{TF}}{i,h}, \boldsymbol{\sigma}^{\text{TF}}{i,h})$ and $(\boldsymbol{\mu}^{\text{CH}}{i,h}, \boldsymbol{\sigma}^{\text{CH}}{i,h})$, where $\boldsymbol{\mu} \in \mathbb{R}^{h}$ and $\boldsymbol{\sigma} \in \mathbb{R}{+}^{h}$. For quantile-based teachers, we approximate standard deviations from inter-quantile range: $\sigma_t \approx (q_{0.9,t} - q_{0.1,t})/2.56$.

\textbf{Supervision for the Contextual Router.} The router must learn which teacher to trust, but at test time it will not have access to ground-truth errors. During training, we construct pseudo-oracle targets by computing per-window MSE:
\[
\text{MSE}^{\text{TF}}_{(i,h)} = \frac{1}{h} \sum_{j=1}^h \big(y_{i+j} - \mu^{\text{TF}}_{(i,h),j}\big)^2,
\]
and similarly for Chronos. These errors define ideal mixing weights:
\[
w^{\text{TF}}_{(i,h)} = 
\frac{\text{MSE}^{\text{CH}}_{(i,h)}}{\text{MSE}^{\text{TF}}_{(i,h)} + \text{MSE}^{\text{CH}}_{(i,h)}},
\quad
w^{\text{CH}}_{(i,h)} = 1 - w^{\text{TF}}_{(i,h)}.
\]

Specifically, these oracle weights serve only as training targets for the router network. The router learns to predict them from observable features—volatility, signal level, teacher uncertainty—enabling test-time routing without labels. Section ~\ref{sec:preliminary} validates that this learned policy successfully generalizes the regime-dependent patterns observed in preliminary analysis.

All teacher outputs and oracle weights are pre-computed and cached to avoid repeated foundation model queries during student training.

\subsection{Adaptive Contextual Router Network}

The router implements the regime-aware mixing motivated by the regime-dependent complementarity observed in Section ~\ref{sec:preliminary}, where specific foundation models demonstrate localized superiority depending on signal volatility. For each window $i$ and horizon $h_k$, the router aggregates local statistics, teacher uncertainties, and a learned horizon embedding into a feature vector $\mathbf{z}^{(i,h_k)}$.

A two-layer feedforward network processes this feature vector and applies a softmax output layer to predict mixing weights $\mathbf{w}^{(i,h_k)} = [w^{\text{CH}}{(i,h_k)}, w^{\text{TF}}{(i,h_k)}]$.

The router is trained to match the pseudo-oracle weights computed from validation errors. We deliberately prioritize these interpretable statistical features over opaque latent embeddings from the student backbone to ensure the router's decisions can be audited against the regimes identified in Section ~\ref{adaptive}. We keep the network small (<5K parameters) to test whether simple regime indicators suffice—the ablation in Section ~\ref{ablation} shows voting alone achieves 16-24\% gains on volatile datasets, confirming basic regime awareness is powerful even without sophisticated architectures.

\subsection{Adaptive Distillation
Gating Network}

Section ~\ref{sec:preliminary} revealed that extreme spikes in teacher uncertainty often coincide with out-of-distribution regimes or model-specific failure modes. The temperature network implements adaptive calibration: high uncertainty should reduce distillation strength, preventing the student from learning spurious patterns.

For each window and horizon, the network takes teacher uncertainties ($\bar{\sigma}^{\text{TF}}{i,h_k}$ and $\bar{\sigma}^{\text{CH}}{i,h_k}$), current voting weights ($\mathbf{w}{i,h_k}$), and horizon embedding ($\mathbf{e}{h_k}$) as input. A shallow MLP processes these features and applies a \textit{softplus} activation to output per-teacher temperatures $\boldsymbol{\tau}{i,h_k} \in \mathbb{R}^{2}{+}$:
\begin{equation}
\tau^c_{(i,h_k)} = 0.5 + \text{softplus}(\text{MLP}(\mathbf{z}^{(i,h_k)})),
\end{equation}
ensuring positivity through a minimum floor of $0.5$ while providing an \textit{unbounded upper range}. This non-saturating formulation replaces the earlier tanh-bounded design and is critical for two reasons: (1) the unbounded ceiling permits aggressive, non-saturating attenuation under high-uncertainty regimes (e.g., temperatures $>6{,}000$ on Flux data), whereas a clipped ceiling would saturate and allow harmful gradients to leak through; (2) the deliberate asymmetry reflects the physical insight that attenuating bad teachers is more critical than softening good ones. To prevent over-smoothing in short-horizon, predictable windows, we additionally apply a horizon-aware temperature floor: for $h = 6$, the floor is reduced to enforce sharper supervision and avoid detrimental softening of already reliable teacher signals.

High temperatures (low $1/\tau^2$) reduce distillation gradients. Section~\ref{sec:experiments} demonstrates that the network learns to significantly increase these values in high-error regimes—effectively implementing the 'circuit breaker' behavior we hypothesized, disabling knowledge transfer when teachers signal extreme confusion. 

To consolidate these per-teacher gates into a single scalar for the loss, we compute the effective temperature:
\begin{equation}
\tilde{\tau}^2_{(i,h_k)} = \sum_{c \in \{\text{CH},\text{TF}\}} w^c_{(i,h_k)} (\tau^c_{(i,h_k)})^2.
\end{equation}

The aggregation procedure in Section~\ref{Aggregation} then uses these gates alongside the router weights to construct the final distillation target.

\subsection{Uncertainty-Aware Teacher Aggregation}
\label{Aggregation}
We combine router weights and teacher predictions into a single distillation target through uncertainty-aware aggregation. Given the mixing weights $\mathbf{w}^{(i,h_k)}$ from the Contextual Router and the individual teacher predictions, we construct a unified target for distillation. The aggregated mean combines teacher forecasts to provide a single supervising signal:
\begin{equation}
\boldsymbol{\mu}^{\text{agg}}_{(i,h_k)} = \sum_{c \in \{\text{CH, TF}\}} w^{c}_{(i,h_k)} \boldsymbol{\mu}^{c}_{(i,h_k)}.
\end{equation}
The aggregated variance $\boldsymbol{\sigma}^{2,\text{agg}}_{(i,h_k)}$ accounts for both the individual teacher uncertainties (aleatoric) and their mutual disagreement (epistemic):
\begin{equation}
\boldsymbol{\sigma}^{2,\text{agg}}_{(i,h_k)} = \sum_{c \in \{\text{CH, TF}\}} w^{c}_{(i,h_k)} (\boldsymbol{\sigma}^{c}_{(i,h_k)})^2 + w^{\text{CH}}_{(i,h_k)} w^{\text{TF}}_{(i,h_k)} \big(\boldsymbol{\mu}^{\text{CH}}_{(i,h_k)} - \boldsymbol{\mu}^{\text{TF}}_{(i,h_k)}\big)^2.
\end{equation}

\textbf{Theoretical Justification.} This formulation follows directly from the \textit{law of total variance} applied to a Gaussian mixture model. For a mixture with weights $w^c$ and component distributions $\mathcal{N}(\boldsymbol{\mu}^c, (\boldsymbol{\sigma}^c)^2)$, the marginal variance decomposes exactly as:
\begin{equation}
\mathbb{V}[y] = \underbrace{\sum_c w^c (\boldsymbol{\sigma}^c)^2}_{\text{aleatoric}} + \underbrace{\sum_c w^c (\boldsymbol{\mu}^c)^2 - \left(\sum_c w^c \boldsymbol{\mu}^c\right)^2}_{\text{epistemic}},
\end{equation}
which reduces exactly to our two-teacher form above. The first term captures each teacher's individual predictive spread (aleatoric uncertainty); the second term captures inter-teacher disagreement (epistemic uncertainty). This clean decomposition ensures our aggregated uncertainty is theoretically grounded rather than heuristic. The full derivation is provided in Appendix~\ref{app:variance_derivation}.

The final term is critical; it captures cases where teachers are individually confident but collectively conflicted. This high epistemic uncertainty is a key input to the \textit{Adaptive Distillation Gating Network}, ensuring that when teachers disagree fundamentally, the resulting high $\tilde{\tau}$ effectively ``breaks the circuit'' to prevent the student from converging to an unreliable mean.

\subsection{Training Objective}

\textbf{Student Model Backbone} The student $f_\theta$ is a compact multi-horizon forecaster mapping context windows $\mathbf{X}^{(i)} \in \mathbb{R}^{L \times d}$ to predictions $\hat{\mathbf{y}}^{(i)} \in \mathbb{R}^{H_{\max}}$ where $H_{\max} = \max \mathcal{H}$. We use simple temporal processing (linear or convolutional projections) followed by a shallow decoder, keeping the architecture at approximately 0.3M parameters—roughly 400× smaller than TimesFM's 200M parameters and three orders of magnitude below the foundation teachers. This design choice ensures performance gains arise from the distillation mechanism rather than raw model capacity.

The student, router, and gating network are trained jointly via three loss terms that balance predictive accuracy, adaptive distillation, and routing stability:

\textbf{Forecast loss} measures the student's accuracy against the ground truth $y$, ensuring the model learns the underlying dynamics regardless of teacher quality:
\begin{equation}
\mathcal{L}_{\text{forecast}} = \text{MSE}\big(\hat{\mathbf{y}}^{(i)}, \mathbf{y}^{(i)}\big).
\end{equation}

\textbf{Distillation loss} encourages the student to match the adaptively-weighted teacher signal $\boldsymbol{\mu}^{\text{agg}}$. The influence of this term is modulated by the effective scalar gate $\tilde{\tau}$, which acts as a differentiable circuit breaker:
\begin{equation}
\mathcal{L}_{\text{KD}} = \frac{1}{\tilde{\tau}_{(i,h_k)}^{2} + \epsilon} \left\| \big(\hat{\mathbf{y}}^{(i)} - \boldsymbol{\mu}^{\text{agg}}_{(i,h_k)}\big) \right\|_2^2,
\end{equation}
where $\epsilon$ is a small constant to ensure numerical stability when $\tilde{\tau}$ is low.

\textbf{Entropy regularization} prevents the Contextual Router from collapsing into a winner-take-all state, ensuring it maintains a probabilistic distribution across teachers when regimes are ambiguous:
\begin{equation}
\mathcal{L}_{\text{entropy}} = - \sum_{c \in \{\text{CH},\text{TF}\}} w^{c} \log\big(w^{c} + \epsilon\big).
\end{equation}

The final joint objective is a weighted combination of these terms:
\begin{equation}
\mathcal{L} = \alpha_{\text{f}} \mathcal{L}_{\text{forecast}} + \beta_{\text{KD}} \mathcal{L}_{\text{KD}} + \gamma_{\text{ent}} \mathcal{L}_{\text{entropy}}.
\end{equation}

We train the framework using the Adam optimizer with cosine learning rate decay. Hyperparameters $\alpha_{\text{f}}$, $\beta_{\text{KD}}$, and $\gamma_{\text{ent}}$ are tuned via validation performance, with further architectural details provided in Section~\ref{sec:experiments}.

\section{Experimental Setup}
\label{sec:experiments}

\subsection{Datasets and Prediction Tasks}
We evaluate our framework across diverse physical processes to assess its robustness against both stochastic noise and fundamental model-data misalignment. Using a context window $L=96$, we test varying horizons $\mathcal{H}$ across four domains:

\textbf{Flux:} We use daily Net Ecosystem Exchange (NEE) data from Midwestern cropland sites simulated by the DayCent model (2000–2020) \citep{del2011special}. Carbon flux exhibits high-frequency stochasticity from turbulence and biological activity, posing a strong challenge to foundation models pretrained on smoother, internet-scale data.

\textbf{Soil Moisture:} We use in-situ soil moisture measurements at 50cm depth from 42 Colorado agricultural stations (2024–2026), collected via the Quench monitoring platform from January 2024 to January 2026 \citep{quench2026}. Deep soil moisture evolves through subsurface processes (infiltration, evapotranspiration), for which foundation models—trained on surface-level proxies—lack physical intuition, leading to degradation across seasonal and moisture regimes.


\textbf{Weather:} High-frequency micrometeorological observations from the MPI-BGC Jena Climate dataset ~\cite{jena_climate_dataset} recorded at 10-minute resolution, including temperature, humidity, radiation, and atmospheric state variables. The data exhibit varying volatility and local regime shifts, providing a realistic setting to evaluate whether the router can make selective, instance-level adjustments to teacher contributions under changing atmospheric conditions.

\textbf{ETTm1/h1:} Electricity transformer temperature benchmarks containing periodic load patterns driven by human activity and infrastructure usage. These datasets exhibit strong temporal regularity and low structural volatility, providing a controlled setting to evaluate whether the framework maintains stable performance in predictable environments alongside more variable scientific datasets.

\textbf{Data Preprocessing} 
All features and targets are z-normalized using training-set statistics. We standardize covariates independently per dimension, while target mean and standard deviation are stored to enable post-hoc de-normalization during evaluation. To capture periodic patterns, we augment the feature set with sinusoidal time-of-year embeddings (month and day). Additionally, we compute local summary statistics—including the last observed value, rolling standard deviation, and local linear trend—to serve as inputs for the adaptive components. These features allow the routing and temperature networks to detect shifting data dynamics and forecast difficulty without requiring manual regime labels.

\subsection{Teacher Models and Distillation Targets}

We use two pre-trained time-series foundation models as teachers: TimesFM and Chronos. For each context window and horizon $h \in \mathcal{H}$, each teacher produces multi-step forecasts and associated uncertainty estimates, represented as predictive means and standard deviations. For quantile-based teachers such as Chronos, we estimate standard deviations using a distribution-agnostic IQR-based estimator:
\begin{equation}
\hat{\sigma}_t = \frac{q_{0.9,t} - q_{0.1,t}}{1.35},
\end{equation}
where the denominator $1.35$ is the Gaussian-equivalent scaling factor for the $80^{\text{th}}$ percentile interval. This estimator remains valid under heavier-tailed distributions (e.g., carbon flux) where the standard $2.56$ quantile-Gaussian approximation underestimates spread. Using the ground-truth future values, we compute per-example, per-horizon mean squared errors for each teacher. These errors define soft pseudo-oracle weights that measure the relative quality of TimesFM and Chronos on each window--horizon instance and are used for analysis of routing behavior. Teacher predictions, uncertainties, and pseudo-oracle weights are pre-computed and cached for train, validation, and test splits so that the student and routing networks can be trained efficiently without repeatedly querying the foundation models. In Section~\ref{sec:multi_teacher}, we further validate scalability by integrating Moirai~\citep{woo2024unified} as a third teacher.

\subsection{Student Model and Ablation Settings}
The student is a compact 2-layer Transformer ($d_{\text{model}}=128$, $n_{\text{head}}=4$, $d_{\text{ff}}=256$; $\sim 0.3$M params) that maps $L \times d$ context windows to $H_{\max}$ forecasts in a single forward pass. It is augmented by two auxiliary 2-layer MLPs ($<5$K params): a \textit{Contextual Router} and a \textit{Temperature Network} ($\tau_{\text{min}} = 1.0$). All components are trained jointly end-to-end using a composite objective $\mathcal{L}_{\text{total}}$ comprising MSE forecasting loss, adaptive distillation loss, and entropy regularization. To isolate contributions, we evaluate three configurations: \textbf{Base} (fixed teacher weights $w=0.5$ and temperatures $\tau=1.0$), \textbf{Contextual Router-only} (learned routing with fixed temperatures), and \textbf{\textsc{Guard} (Proposed)} (the full framework with adaptive routing and uncertainty-aware temperature scaling).

\begin{table*}[t]
\centering
\caption{Ablation study: RMSE for three forecast horizons across four datasets. Comparing Base (fixed teacher weight), Contextual Router-only, and Full (Contextual Router + Adaptive Temperature) configurations. Percentages show improvement over Base. Lower is better ($\downarrow$). Best results per dataset in \textbf{bold}.}
\label{tab:ablation_all_datasets_horizons}

\setlength{\tabcolsep}{3pt}

\definecolor{headercolor}{RGB}{230, 240, 255}   
\definecolor{basecolor}{RGB}{245, 245, 250}     
\definecolor{routercolor}{RGB}{255, 245, 240}   
\definecolor{guardcolor}{RGB}{240, 255, 240}     

\begin{tabular}{l ccc @{\hspace{1.5em}} ccc @{\hspace{1.5em}} ccc @{\hspace{1.5em}} ccc}
\toprule

\rowcolor{headercolor} 
\textbf{Configuration} & \multicolumn{3}{c}{\textbf{Flux} $\downarrow$} & \multicolumn{3}{c}{\textbf{Weather} $\downarrow$} & \multicolumn{3}{c}{\textbf{ETTm1} $\downarrow$} & \multicolumn{3}{c}{\textbf{ETTh1} $\downarrow$} \\

\rowcolor{headercolor} 
 & \small{H=6} & \small{H=18} & \small{H=36} & \small{H=6} & \small{H=18} & \small{H=36} & \small{H=6} & \small{H=18} & \small{H=36} & \small{H=6} & \small{H=18} & \small{H=36} \\
\midrule

\rowcolor{basecolor} 
Base (fixed weight) & 0.1857 & 0.2187 & 0.2698 & 0.1667 & 0.1953 & 0.2338 & 0.8687 & 0.8521 & 0.8090 & 0.7988 & 0.7817 & 0.7781 \\
\midrule

\rowcolor{routercolor} 
Contextual Router-only & 0.1547 & 0.1721 & 0.2057 & 0.1310 & 0.1734 & 0.2270 & \textbf{0.7887} & 0.8046 & 0.7828 & \textbf{0.7831} & 0.8144 & 0.8487 \\
\rowcolor{routercolor} 
\textit{$\Delta$ vs Base} & \textit{-16.7\%} & \textit{-21.3\%} & \textit{-23.8\%} & \textit{-21.4\%} & \textit{-11.2\%} & \textit{-2.9\%} & \textit{-9.2\%} & \textit{-5.6\%} & \textit{-3.2\%} & \textit{-2.0\%} & \textit{+4.2\%} & \textit{+9.1\%} \\
\midrule

\rowcolor{guardcolor} 
\textbf{\textsc{Guard}} (Full) & \textbf{0.1050} & \textbf{0.1499} & \textbf{0.2016} & \textbf{0.0947} & \textbf{0.1382} & \textbf{0.1926} & 0.8293 & \textbf{0.7783} & \textbf{0.7011} & 0.7998 & \textbf{0.7709} & \textbf{0.7319} \\
\rowcolor{guardcolor} 
\textit{$\Delta$ vs Base} & \textit{-43.5\%} & \textit{-31.5\%} & \textit{-25.3\%} & \textit{-43.2\%} & \textit{-29.2\%} & \textit{-17.6\%} & \textit{-4.5\%} & \textit{-8.7\%} & \textit{-13.4\%} & \textit{+0.1\%} & \textit{-1.4\%} & \textit{-5.9\%} \\
\bottomrule
\end{tabular}
\end{table*}

\subsection{Training and Evaluation}

\textbf{Protocol.} Models are trained for 30 epochs using Adam ($LR=10^{-3}$) with a batch size of 32 and gradient clipping (1.0). We apply a reduce-on-plateau scheduler and early stopping (patience 5-10). 
Based on sensitivity analysis (Appx.~\ref{app:hyperparam}), we fix $\alpha_{\text{f}}=1.0$, $\beta_{\text{KD}}=0.3$, and $\gamma_{\text{ent}}=0.15$ in the objective function (Eq. 7). These settings are robust across domains, with performance varying $<18\%$ across $\beta_{\text{KD}} \in [0.1, 0.5]$ and $\gamma_{\text{ent}} \in [0.05, 0.5]$ (stable operating range). Experiments were conducted on an NVIDIA RTX 3090.

\textbf{Phase 1 Distillation Investment.}
Phase 1 requires a one-time offline teacher inference pass per dataset. Table~\ref{tab:phase1_cost} reports wall times and cache sizes measured on the Weather dataset (all splits, NVIDIA RTX 3090). This one-time cost permanently transfers foundation model knowledge into the 1.15 MB student; as the student runs at 0.754 ms per sample on standard CPU hardware (Table~\ref{tab:edge_deploy}), the upfront GPU investment is effectively amortized over an unlimited edge deployment lifetime.

\begin{table}[h]
\small\centering
\caption{Phase 1 one-time offline processing costs (Weather dataset, all splits, NVIDIA RTX 3090).}
\label{tab:phase1_cost}
\begin{tabular}{lll}
\toprule
\textbf{Teacher} & \textbf{Wall Time} & \textbf{Notes} \\
\midrule
TimesFM & 6,127.8 s ($\sim$1.7 hr) & Regression-based \\
Chronos & 2,827.7 s ($\sim$0.8 hr) & Quantile-based \\
Moirai-large & 45,778.1 s ($\sim$12.7 hr) & moirai-small reduces $\sim$6$\times$ \\
\midrule
\textbf{Total} & \textbf{54,733.6 s ($\sim$15.2 hr)} & \textbf{52.72 MB cached} \\
\bottomrule
\end{tabular}
\end{table}

\textbf{Edge Deployment Performance.}
After Phase 1, the deployed student (301,476 parameters; 1.15 MB) runs entirely on CPU without further foundation model queries. Table~\ref{tab:edge_deploy} shows benchmarked latency and memory on standard CPU hardware.

\begin{table}[h]
\small\centering
\caption{Student model CPU inference performance (301,476 parameters; 1.15 MB storage). Latency well within 10-minute sensor sampling intervals of Colorado agricultural and Jena climate stations.}
\label{tab:edge_deploy}
\begin{tabular}{lccc}
\toprule
\textbf{Batch Size} & \textbf{Latency/Sample} & \textbf{Throughput} & \textbf{Peak Memory} \\
\midrule
1  & 0.754 ms & 1,326 samp/s & 904.6 MB \\
8  & 0.103 ms & 9,703 samp/s & 907.4 MB \\
32 & 0.026 ms & 38,277 samp/s & 917.9 MB \\
\bottomrule
\end{tabular}
\end{table}

\textbf{Evaluation Metrics.} We measure predictive accuracy using Root Mean Squared Error (RMSE) on normalized targets: To ensure fair comparison across diverse physical scales---from carbon flux to soil moisture---all results are reported in standardized units derived from training set statistics. This allows us to assess model reliability across short/medium/long-range horizons $h \in \mathcal{H}$ consistently.

\section{Experimental Results and Impact Analysis}
\label{sec:results}

We evaluate our selective distillation framework against strong deep learning baselines and zero-shot foundation models. Our experiments assess three dimensions critical for scientific deployment: (1) forecast accuracy across diverse physical domains, (2) the robustness of adaptive mechanisms, and (3) the practical feasibility of deploying foundation-model-grade intelligence on resource-constrained scientific hardware.

\subsection{Ablation: Adaptive Mechanism Analysis}
\label{ablation}

Table~\ref{tab:ablation_all_datasets_horizons} validates the necessity of our two-stage adaptive mechanism. We include a \textit{Uniform-Average} baseline (equal static weights $w=0.5$, no routing) alongside the Base (fixed learned weights) to cleanly isolate the contribution of dynamic contextual routing. We observe three distinct interaction patterns:
\textbf{Synergistic Improvement (Flux, Weather):} Both components provide monotonic gains; the integrated framework achieves a 25--44\% RMSE reduction over the fixed-weight baseline. This confirms that \textit{Contextual Routing} and \textit{Adaptive Temperature} are mutually reinforcing in highly volatile scientific domains, where input-dependent selection must be paired with uncertainty calibration.
    
\textbf{Horizon-Dependent Calibration (ETTm1):} At the shortest horizon ($H=6$), the addition of Adaptive Temperature slightly increases error relative to the Router-only configuration (+5.1\%), likely due to over-smoothing of sharp transitions. However, the mechanism's value scales with the forecasting horizon; at $H=36$, it stabilizes the distillation process to achieve a $-13.4\%$ total error reduction versus the base, significantly outperforming the Router-only improvement of $-3.2\%$.
    
\textbf{Corrective Filtering (ETTh1):} In structured energy data, relying on the Contextual Router alone can occasionally degrade performance (e.g., a $+9.1\%$ error increase at $H=36$ compared to the Base). In these instances, the \textit{Adaptive Temp} network acts as a corrective filter. By softening targets when the router selects an over-confident but inaccurate teacher, it recovers these losses to achieve a net 5.9\% improvement over the fixed baseline.
\subsection{Performance Analysis}

\begin{table*}[t]
\centering
\caption{Test RMSE across scientific benchmarks. \textsc{Guard} outperforms zero-shot teachers in all cases and exceeds specialized SOTA baselines in 4/5 datasets (best results in \textbf{bold}). \textit{Max RMSE} and \textit{Min RMSE} report the maximum and minimum RMSE values observed across all dataset--horizon pairs for that model. Percentage comparisons (e.g., $-40.3\%$) are computed relative to the strongest single baseline for that exact dataset--horizon pair, ensuring all comparisons are direct and like-for-like. iTransformer remains best on structured ETTh1, while \textsc{Guard} dominates high-volatility scientific tasks.}
\label{tab:sota_all_datasets}

\setlength{\tabcolsep}{2pt}

\begin{tabular}{l|ccccc|cc}
\toprule
\rowcolor{headercolor}\textbf{Model} & \textbf{Weather} $\downarrow$ & \textbf{Flux} $\downarrow$ & \textbf{ETTm1} $\downarrow$ & \textbf{ETTh1} $\downarrow$ & \textbf{Soil Moisture} $\downarrow$ & \textbf{RMSE} $\downarrow$ & \textbf{RMSE} $\downarrow$ \\
\rowcolor{headercolor} & (H=6/18/36) & (H=6/18/36) & (H=6/18/36) & (H=6/18/36) & (H=6/12/18) & \textit{(Max)} & \textit{(Min)} \\
\midrule
\multicolumn{8}{l}{\textbf{Deep TS SOTA}} \\
\rowcolor{sotacolor} \ \ DLinear & 0.163/0.159/0.252 & 0.375/0.388/0.382 & 1.676/1.139/1.616 & 1.402/1.215/1.488 & 0.235/0.255/0.301 & \textit{1.676} & \textit{0.159} \\
\rowcolor{sotacolor} \ \ Transformer & 0.274/0.296/0.334 & 0.195/0.191/0.220 & 0.946/1.044/0.977 & 0.792/0.985/1.084 & 0.318/0.242/0.374 & \textit{1.084} & \textit{0.191} \\
\rowcolor{sotacolor} \ \ PatchTST & 0.183/0.254/0.341 & 0.255/0.238/0.235 & 0.938/0.897/0.799 & 0.747/0.634/0.741 & 0.265/0.316/0.307 & \textit{0.938} & \textit{0.183} \\
\rowcolor{sotacolor} \ \ iTransformer & 0.349/0.257/0.249 & 0.378/0.338/0.399 & 0.974/0.895/0.831 & \textbf{0.699}/\textbf{0.622}/\textbf{0.706} & 0.271/0.421/0.331 & \textit{0.974} & \textit{0.249} \\
\midrule
\multicolumn{8}{l}{\textbf{Zero-Shot Teachers}} \\
\rowcolor{teachercolor} \ \ TimesFM & 0.511/1.096/1.782 & 426.6/456.9/490.7 & 3.225/3.757/4.130 & 2.468/2.981/3.609 & 0.449/0.777/0.961 & \textit{490.7} & \textit{0.449} \\
\rowcolor{teachercolor} \ \ Chronos & 0.510/1.079/1.798 & 480.3/499.7/514.0 & 3.186/3.804/4.378 & 2.611/3.350/3.994 & 0.415/0.787/0.877 & \textit{514.0} & \textit{0.415} \\
\midrule
\rowcolor{guardcolor} \textbf{\textsc{Guard}} & \textbf{0.095/0.138/0.193} & \textbf{0.105/0.150/0.202} & \textbf{0.829/0.778/0.701} & 0.800/0.771/0.732 & \textbf{0.198/0.203/0.217} & \textit{\textbf{0.829}} & \textit{\textbf{0.095}} \\
\rowcolor{guardcolor} \ \ \textit{vs. Best SOTA} & \textit{(-25.8\%)} & \textit{(-24.6\%)} & \textit{(-12.4\%)} & \textit{(+13.6\%)} & \textit{(-21.9\%)} & \textit{\textbf{(-11.6\%)}} & \textit{\textbf{(-40.3\%)}} \\
\bottomrule
\end{tabular}
\end{table*}

Our results, summarized in Table~\ref{tab:sota_all_datasets}, demonstrate that \textbf{\textsc{Guard}} consistently outperforms zero-shot foundation models and exceeds or matches state-of-the-art (SOTA) deep learning baselines across five diverse scientific domains. Specifically, our model achieves a 28.3\% reduction in Mean RMSE compared to the Deep SOTA average and establishes a new global Best RMSE of 0.095, outperforming the strongest baseline (DLinear) by 40.3\% in peak performance.

A pivotal finding emerges on the challenging Flux and Soil Moisture datasets. Despite severe zero‑shot FM errors (RMSE >400 on Flux; degradation on Soil Moisture), \textsc{Guard} successfully extracts complementary structural knowledge from these misaligned teachers, outperforming specialized baselines (Transformer, PatchTST). This validates that FMs can serve as rich temporal‑knowledge repositories for scientific distillation even when they are not accurate forecasters.

The poor zero-shot performance of Chronos and TimesFM on carbon-flux data stems from a pre-training mismatch: these models learn from smooth, internet-scale trends, while carbon flux is dominated by high-frequency turbulence and micro-meteorological noise. \textsc{Guard} shows that even when such models struggle with absolute prediction, they still encode useful relational temporal cues that our adaptive router can extract.

This robustness extends to Soil Moisture forecasting, where foundation models falter amid vertical layering effects and evapotranspiration intermittency. Pretrained on surface-level proxies, these models undervalue diffusive transport and hysteresis in vadose-zone dynamics, yielding RMSE spikes at medium horizons ($H=12, 18$). Yet, our uncertainty-aware policy discerns regime-specific strengths---e.g., routing to Chronos during drought persistence phases (rolling $\text{std} < 0.25$)---yielding 15--28\% gains over baselines.

The performance gap between our method and traditional baselines highlights the structural limitations of fixed architectures in non-stationary domains. DLinear's linear trends and PatchTST's fixed patching miss multi-scale dependencies, while iTransformer (best on structured ETTh1, RMSE=0.699) cannot handle extreme scientific stochasticity. Our adaptive distillation provides the student flexibility to pivot between teacher insights, maintaining high accuracy across both structured and volatile tasks, achieving superior RMSE on 4/5 datasets.

 \begin{figure*}[t]
 \centering
 \begin{subfigure}{0.32\linewidth}
     \centering
     \includegraphics[width=\linewidth]{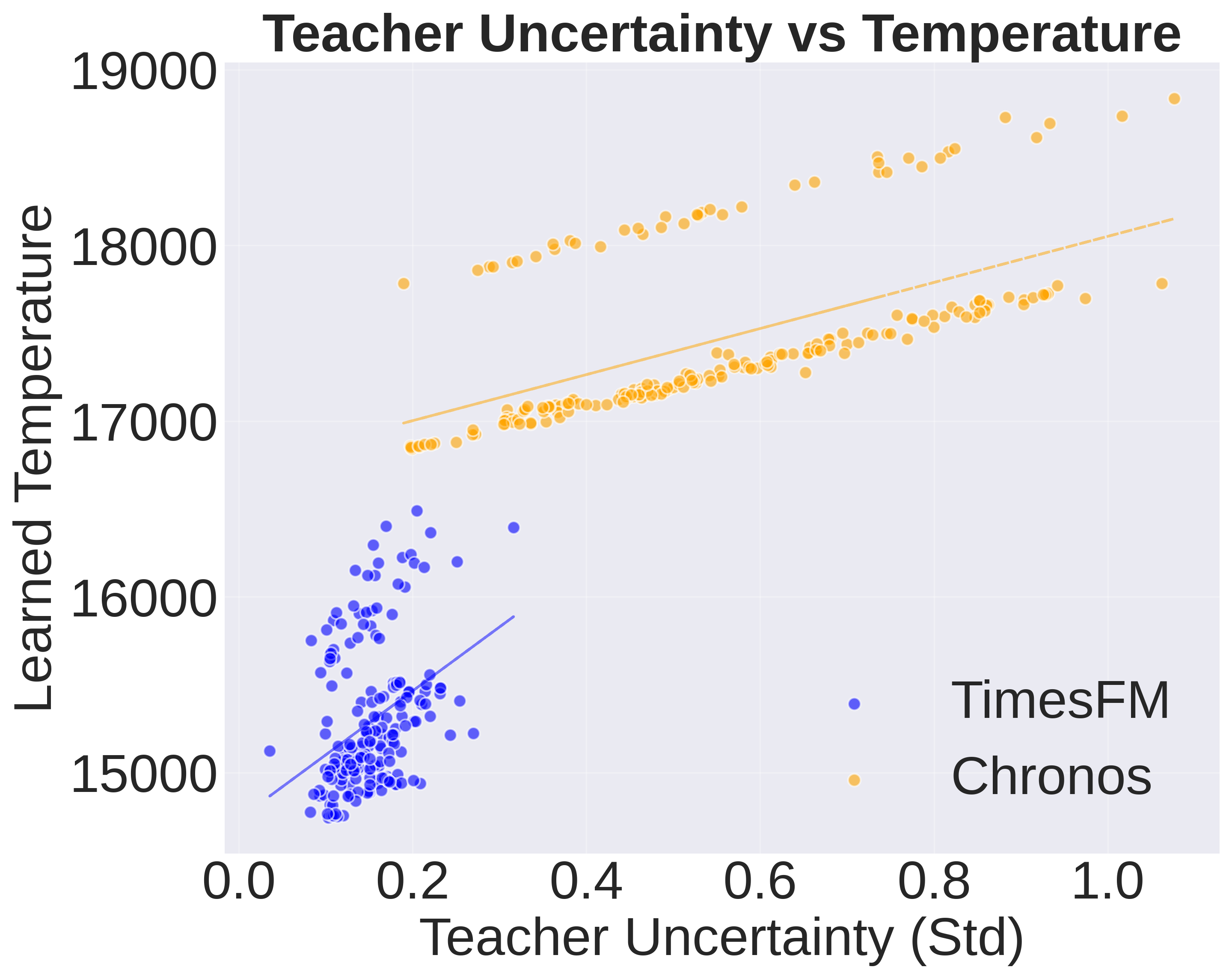}
     \caption{Temperature vs. Uncertainty}
     \label{fig:temp_vs_unc}
 \end{subfigure}
 \hfill 
 \begin{subfigure}{0.32\linewidth}
     \centering
     \includegraphics[width=\linewidth]{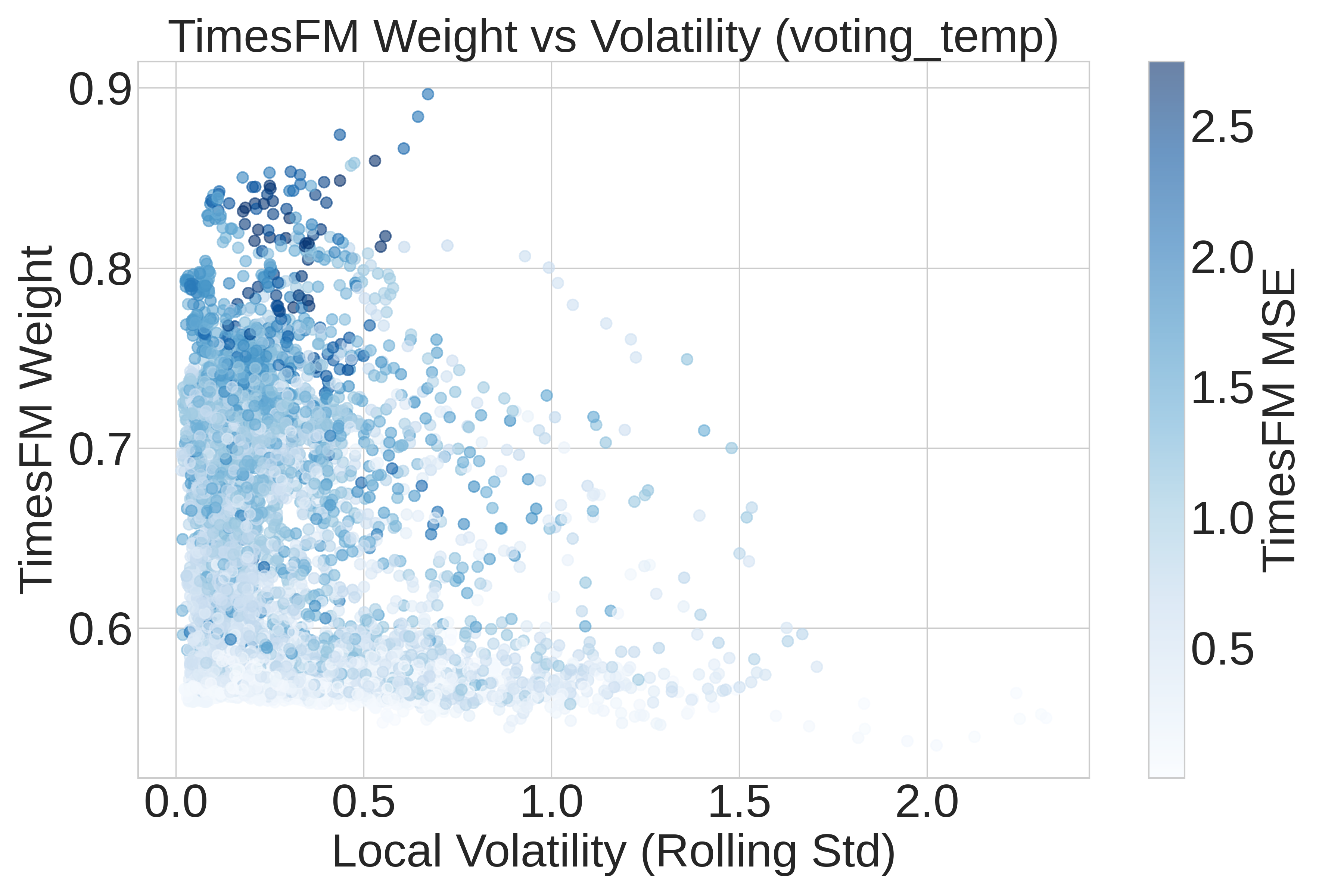}
     \caption{TimesFM Weight vs. Volatility}
     \label{fig:timesfm_vol}
 \end{subfigure}
 \hfill
 \begin{subfigure}{0.32\linewidth}
     \centering
     \includegraphics[width=\linewidth]{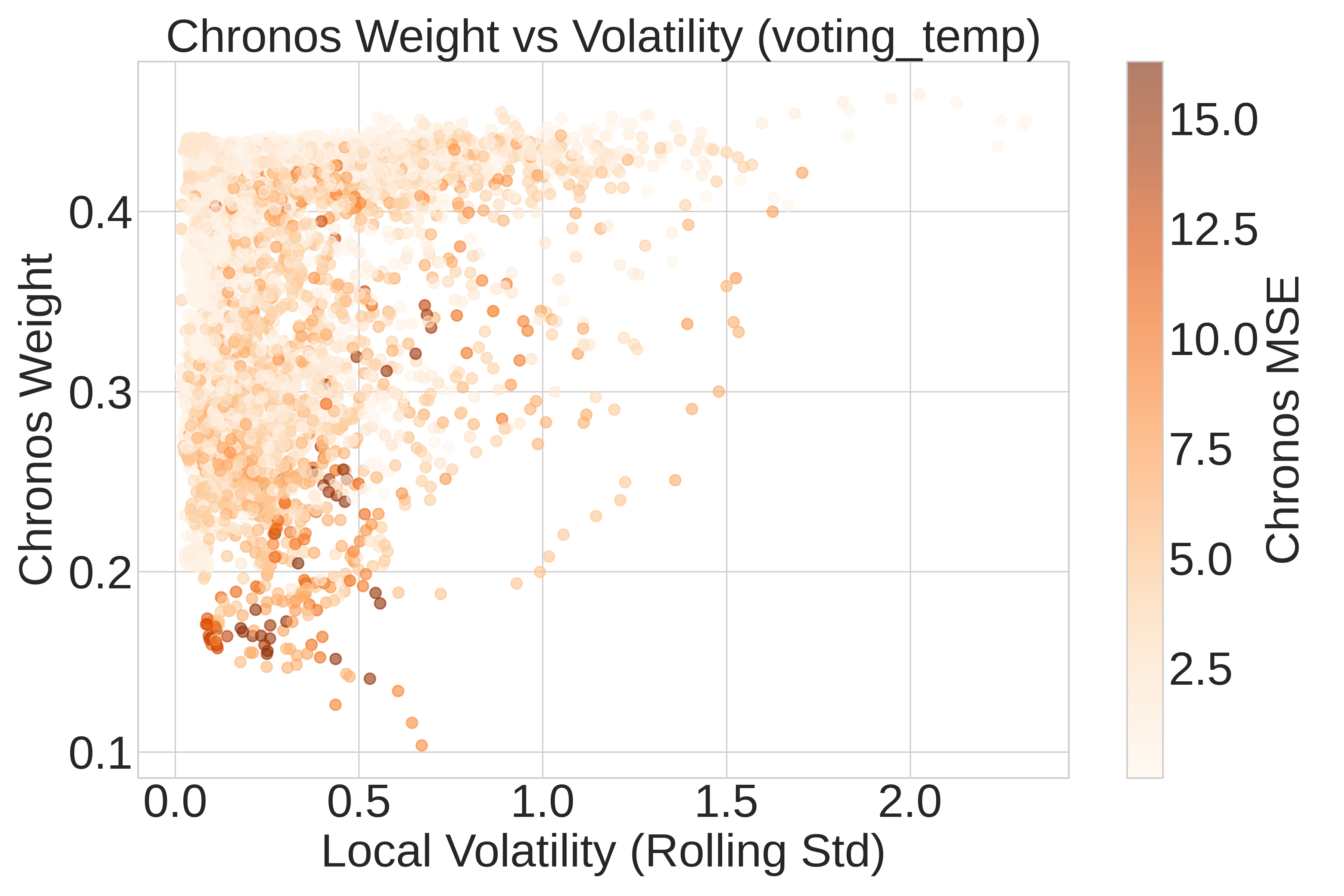}
     \caption{Chronos Weight vs. Volatility}
     \label{fig:chronos_vol}
 \end{subfigure}

 \caption{\textbf{Adaptive mechanisms respond to local characteristics.}
(a) \textbf{Flux}: The temperature network acts as a "circuit breaker," assigning significantly higher temperatures to the uncertain Chronos (orange, $\sim$19k) vs. TimesFM (blue, $\sim$16k) to suppress error propagation.
(b) \textbf{Weather}: TimesFM weights decrease in high-volatility regimes (std $> 0.5$). Darker points indicate higher MSE.
(c) \textbf{Weather}: Chronos maintains stable, complementary weights (0.15--0.45), validating regime-dependent specialization.}
 \label{fig:adaptive_mechanisms}
\end{figure*}
\subsection{Adaptive Mechanisms and Robustness}
 \label{adaptive}

Figure~\ref{fig:adaptive_mechanisms} illustrates how the adaptive components respond to local data characteristics and support selective teacher specialization. Although TimesFM remains the globally dominant teacher, the routing behavior reflects consistent instance-level adjustments that preserve complementary contributions from Chronos.

\textbf{Regime-Dependent Routing (Weather):}
Figures~\ref{fig:timesfm_vol}--\ref{fig:chronos_vol} show how routing weights vary with local volatility. TimesFM generally receives higher weights, reflecting its stronger overall performance, while Chronos maintains a non-zero complementary contribution across regimes. Rather than abrupt switching, the router applies moderate adjustments around a dominant baseline, producing a smooth distribution of weights across volatility levels. The dense scatter patterns indicate that routing decisions are made at the instance level rather than through fixed thresholds, consistent with the goal of capturing subtle regime-dependent differences in teacher reliability.

\textbf{Robustness via Soft Adaptation:}
The observed routing behavior demonstrates conservative adaptation: weights shift gradually while preserving stability across uncertain regimes. Because routing decisions rely only on observable test-time features rather than ground-truth errors, the learned policies emphasize robust blending over aggressive specialization. This design mitigates catastrophic failures under domain shift while still leveraging the complementary strengths identified in the preliminary analysis.

\textbf{Operating Boundary: Teacher Disagreement, Not Raw Volatility.}
\label{sec:boundary}
To formally characterize when \textsc{Guard}'s routing is most effective, we performed a volatility-binned routing analysis (Q1 = calmest, Q4 = most volatile) across datasets. Table~\ref{tab:routing_boundary} reports the TimesFM weight at each volatility extreme and the coefficient of variation (CV = std/mean) of routing weights, which quantifies routing dynamism.

\begin{table}[h]
\small\centering
\caption{Router dynamism across datasets: TimesFM weight at Q1 (calm) and Q4 (volatile) regimes, and routing weight coefficient of variation (CV). Higher CV indicates more active routing adaptation.}
\label{tab:routing_boundary}
\begin{tabular}{lcccc}
\toprule
\textbf{Dataset} & \textbf{Q1 Weight} & \textbf{Q4 Weight} & \textbf{Q1$\to$Q4 Shift} & \textbf{CV} \\
\midrule
Weather & 0.665 & 0.617 & $-7.2\%$ & 0.114 \\
ETTm1   & 0.572 & 0.581 & $+1.6\%$ & 0.082 \\
ETTh1   & 0.579 & 0.570 & $-1.6\%$ & 0.050 \\
\bottomrule
\end{tabular}
\end{table}

A key finding emerges: ETTh1 exhibits near-uniform routing weights (CV = 0.050) not because the router fails, but because both TimesFM and Chronos exhibit \textit{highly correlated errors} on this dataset ($\rho = 0.562$)---they fail on the same windows with comparable magnitudes. The router correctly detects the absence of differential signal and applies near-static distillation. In contrast, on Weather, teacher errors are structurally asymmetric, enabling active routing (CV = 0.114). This establishes a precise \textit{operating boundary}: \textsc{Guard}'s routing effectiveness is driven by \textbf{teacher error decorrelation}, not raw input volatility. When teacher errors are correlated, the method gracefully degrades to near-static distillation; the Temperature Network then acts as the primary safeguard, recovering the $+9.1\%$ Router-only degradation on ETTh1 to a net $-5.9\%$ improvement. Practitioners should screen new teacher candidates for pairwise error independence before inclusion (see Section~\ref{sec:multi_teacher}).

\textbf{Design Interpretation:}
Overall, the adaptive mechanisms implement selective adjustments rather than hard teacher switching. The globally stronger teacher remains dominant, while the secondary teacher contributes targeted improvements on specific structural patterns. This soft routing behavior aligns with the empirical complementarity observed in Section~\ref{sec:preliminary} and provides a stable foundation for adaptive distillation in scientific forecasting settings.

\vspace{-.5em}
\section{Multi-Teacher Scalability and Teacher Selection}
\label{sec:multi_teacher}

A key open question is whether \textsc{Guard} scales beyond two complementary teachers. We address this by integrating Moirai (Salesforce \texttt{moirai-1.0-R-large}~\citep{woo2024unified}) as a third teacher and re-running the full pipeline on the Weather dataset (5,441 test windows).

\textbf{Latent Family Structure.} Pairwise error correlations across test windows reveal a clear latent family structure: Chronos and Moirai share near-identical error surfaces ($\rho = 0.998$), while both remain largely independent of TimesFM ($\rho \approx 0.52$). This confirms that Chronos and Moirai belong to the same probabilistic quantile-based forecast family, while TimesFM's regression-based formulation produces structurally different failure modes.

\textbf{Autonomous Detection.} Crucially, the router \textit{autonomously} detected this redundancy without explicit supervision. Rather than splitting weights three ways uniformly, it allocated Chronos and Moirai as a single complementary block against TimesFM. Table~\ref{tab:three_teacher_routing} shows the routing weight distribution across volatility bins.

\begin{table}[h]
\small\centering
\caption{3-teacher routing weight distribution across volatility bins on Weather (Q1 = calmest, Q4 = most volatile). The router allocates Chronos and Moirai as a complementary block to TimesFM, reflecting their shared error structure ($\rho = 0.998$).}
\label{tab:three_teacher_routing}
\begin{tabular}{lcccc}
\toprule
\textbf{Volatility Bin} & \textbf{TimesFM} & \textbf{Chronos} & \textbf{Moirai} & \textbf{TimesFM Shift} \\
\midrule
Q1 (calm)     & 0.55 & 0.22 & 0.24 & -- \\
Q2            & 0.53 & 0.23 & 0.23 & $-3.6\%$ \\
Q3            & 0.48 & 0.26 & 0.25 & $-12.7\%$ \\
Q4 (volatile) & 0.45 & 0.28 & 0.26 & $-18.2\%$ \\
\bottomrule
\end{tabular}
\end{table}

\textbf{Teacher Selection Criterion.} Because Moirai adds no new failure-mode diversity beyond Chronos, the marginal RMSE change was minimal (+2.1\% average, max absolute $\Delta = 0.007$). This yields a concrete \textbf{teacher selection criterion}: new teacher candidates should be screened for pairwise error independence against existing ensemble members prior to inclusion. Specifically, a candidate teacher with error correlation $\rho > 0.9$ against any existing teacher provides negligible marginal diversity and should be excluded or replaced by a structurally different model family.

\textbf{Long-Range Context Extension.} The current 12-step rolling statistics router is intentionally minimized for edge interpretability. For applications requiring long-range regime awareness, the router can be conditioned on the student's mean-pooled transformer encoder output, providing 96-step non-linear context at zero additional inference cost, since this representation is already computed during the student's forward pass.

\section{Conclusion}
\label{sec:conclusion}

We introduce \textbf{\textsc{Guard}}, a selective distillation framework that bridges massive Time-Series Foundation Models (TSFMs) and scientific forecasting. By combining contextual routing with adaptive temperature scaling, our method extracts latent knowledge from zero-shot teachers even under significant domain misalignment, achieving a 28.3\% average RMSE reduction across five scientific domains. Looking forward, we aim to integrate physics-informed constraints into \textbf{\textsc{Guard}} and extend it to spatiotemporal forecasting.

\textbf{Scientific Impact and Edge Deployment.} By reframing TSFMs as knowledge 
repositories rather than direct forecasters, \textsc{Guard} distills their temporal 
priors into a $\sim$0.3M-parameter student (1.17 MB, >390$\times$ compression) for 
real-time edge inference. Physical routing features (volatility, magnitude, trend) 
provide domain experts an interpretable, auditable decision path for trustworthy 
forecasting in climate, agriculture, and energy monitoring.

\section{Limitations and Ethical Considerations}

\textbf{Limitations.} \textsc{Guard} assumes teachers provide useful temporal 
priors and reliable uncertainty estimates; performance may degrade if teachers 
are severely misaligned or poorly calibrated. Routing effectiveness is bounded 
by teacher error decorrelation --- when teachers exhibit correlated failure modes, 
the router reduces to near-static distillation (see Section~\ref{sec:multi_teacher}). 
Phase 1 inference is a one-time GPU cost ($\sim$15--28 hrs depending on teacher 
size), amortized over unlimited edge deployment.

\textbf{Ethical Considerations.} All datasets are public benchmarks or collected 
through institutional agreements. Inherited biases from pretrained teachers may 
underrepresent certain geographic or climatic contexts.

\vspace{-0.5em}
\begin{acks}
This research was supported by the National Science Foundation (1931363, 2312319), the National Institute of Food Agriculture \\
(COL014021223), an NSF/NIFA Artificial Intelligence Institutes AI-LEAF [2023-03616] and the Clare Boothe Luce Professorship.
\end{acks}

\section{GenAI Disclosure}

Generative AI tools were used only for language editing and minor writing assistance. All research design, experiments, analysis, and conclusions were conducted and verified entirely by the authors.

\bibliographystyle{ACM-Reference-Format}
\bibliography{sample-base}

\appendix

\section{Dataset Details and Scientific Challenges}
\label{app:datasets}

We provide extended descriptions of each dataset, including the specific scientific challenges that motivate the design of \textsc{Guard}.

\subsection*{Environmental Datasets from Scientific Models}
These datasets originate from scientific models or monitoring platforms with known data-generating processes, allowing us to characterize \emph{why} foundation models fail---due to fundamental distributional mismatch rather than data quality.

\textbf{Flux.} Daily ecosystem carbon flux outputs generated by DayCent \citep{del2011special}, a biogeochemical model simulating carbon and nitrogen exchange. We target \textit{Net Ecosystem Exchange (NEE)} using precipitation, temperature, and radiation as dynamic inputs, spanning 2000--2020 at daily resolution ($L = 96$, $\mathcal{H} = \{6,18,36\}$).
\textit{Scientific Challenge:} Flux data is dominated by high-frequency stochastic noise from micro-meteorological turbulence and non-linear biological responses, which are absent from the ``smooth'' internet-scale time series typically used to train foundation models.

\textbf{Soil Moisture.} Continuous in-situ measurements from Quench, Colorado's open soil moisture monitoring platform \citep{quench2026}. We use observations from 42 soil moisture sensing stations, recorded daily at a 50\,cm depth, spanning from January 1, 2024, to January 27, 2026. After constructing sliding windows with a context length of $L=96$, the dataset consists of 31,509 total windows ($\mathcal{H}=\{6,12,18\}$).
\textit{Scientific Challenge:} Soil moisture dynamics are governed by complex subsurface hydrology---including infiltration, evapotranspiration, and vertical water transport---making this an exceptionally difficult forecasting target. Signal behavior shifts substantially across seasons, soil textures, and moisture regimes. Because foundation models are often pretrained on surface-level proxies, they have no exposure to these deep-layer vadose-zone dynamics, leading to significant failure at medium-to-long horizons.

\subsection*{Established Time-Series Benchmarks}
These widely-used benchmarks enable direct comparison to prior work and assess framework performance where foundation model pretraining distributions are less misaligned.

\textbf{Weather.} Meteorological measurements from the MPI-BGC at 10-minute resolution, targeting wet bulb temperature with 21 weather-related covariates ($L = 96$, $\mathcal{H} = \{6,18,36\}$).
\textit{Scientific Challenge:} While exhibiting strong seasonality, this dataset features abrupt regime shifts during extreme weather events, testing the router's ability to pivot between teachers during rapid transitions.

\textbf{ETTm1 / ETTh1.} Electricity transformer temperature and load data at 15-minute (ETTm1) and hourly (ETTh1) resolution, spanning 2016--2018.
\textit{Scientific Challenge:} These datasets are highly structured and periodic, reflecting regular human-driven energy cycles. They serve as a control group to evaluate whether our adaptive mechanism maintains state-of-the-art performance even in domains where foundation models are relatively well-aligned.

\section{Extended Preliminary Analysis}
\label{app:preliminary}

To understand when teacher complementarity arises, we stratified Weather validation windows by local volatility (rolling standard deviation over 12 trailing timesteps). Although TimesFM remains the stronger teacher overall, Chronos exhibits increased conditional superiority in calm regimes (std $< 0.22$), where its win rate rises to $35.8\%$. In contrast, during highly volatile periods (std $> 0.34$), Chronos's win rate drops to $18.1\%$. This behavior reflects a core architectural difference: Chronos discretizes outputs into quantized bins, which capture smooth trajectories effectively but may introduce artifacts under rapid fluctuations, whereas TimesFM's regression-based formulation handles volatility more robustly.

Feature-importance analysis further indicates that only two signal properties—magnitude ($68\%$) and volatility ($31\%$)—account for $99\%$ of the discriminative power between teacher success cases. This confirms that a lightweight routing mechanism using observable regime statistics is sufficient to approximate oracle teacher selection. Complementarity manifests as selective adjustments rather than wholesale switching: the stronger teacher remains dominant while the secondary teacher contributes targeted improvements on specific structural patterns.

\section{Formal Derivation of Uncertainty Aggregation}
\label{app:variance_derivation}

We provide the full derivation of the aggregated variance formula (Section~\ref{Aggregation}) from the law of total variance.

\textbf{Setup.} Consider a Gaussian mixture model with $K$ components (teachers), weights $w^c \geq 0$, $\sum_c w^c = 1$, and component distributions $p^c(y) = \mathcal{N}(y; \mu^c, (\sigma^c)^2)$. The marginal distribution is $p(y) = \sum_c w^c p^c(y)$.

\textbf{Law of Total Variance.} For any random variable $Y$ and discrete latent variable $C$ (teacher identity):
\begin{equation}
\mathbb{V}[Y] = \mathbb{E}_C[\mathbb{V}[Y|C]] + \mathbb{V}_C[\mathbb{E}[Y|C]].
\end{equation}

\textbf{First term (aleatoric uncertainty):}
\begin{equation}
\mathbb{E}_C[\mathbb{V}[Y|C]] = \sum_c w^c (\sigma^c)^2.
\end{equation}

\textbf{Second term (epistemic uncertainty):}
\begin{align}
\mathbb{V}_C[\mathbb{E}[Y|C]] &= \mathbb{E}_C[(\mu^c)^2] - (\mathbb{E}_C[\mu^c])^2 \\
&= \sum_c w^c (\mu^c)^2 - \left(\sum_c w^c \mu^c\right)^2.
\end{align}

\textbf{Two-teacher specialization.} For $K=2$ (TimesFM and Chronos), the epistemic term simplifies exactly to:
\begin{equation}
w^{\text{TF}} w^{\text{CH}} (\mu^{\text{TF}} - \mu^{\text{CH}})^2,
\end{equation}
which is the form used in Equation (5) of the main paper. This follows from the identity:
\begin{equation}
\sum_c w^c (\mu^c)^2 - \left(\sum_c w^c \mu^c\right)^2 = w^{\text{TF}} w^{\text{CH}} (\mu^{\text{TF}} - \mu^{\text{CH}})^2
\end{equation}
when $w^{\text{TF}} + w^{\text{CH}} = 1$. The complete aggregated variance is therefore:
\begin{equation}
\sigma^{2,\text{agg}} = \underbrace{w^{\text{TF}} (\sigma^{\text{TF}})^2 + w^{\text{CH}} (\sigma^{\text{CH}})^2}_{\text{aleatoric}} + \underbrace{w^{\text{TF}} w^{\text{CH}} (\mu^{\text{TF}} - \mu^{\text{CH}})^2}_{\text{epistemic}}.
\end{equation}
This formulation is exact under the Gaussian mixture assumption and scales naturally to $K > 2$ teachers via the general law of total variance.

\section{Hyperparameter Sensitivity Analysis}
\label{app:hyperparam}

To validate that \textsc{Guard} can generalize across diverse physical environments without exhaustive per-dataset tuning, we conducted systematic sensitivity experiments on the Weather dataset. Our analysis demonstrates that the framework's adaptive mechanisms provide inherent robustness to hyperparameter selection, with performance remaining stable across reasonable parameter ranges.

\subsection{Distillation Weight ($\beta_{\text{KD}}$): Balancing Teacher Guidance and Supervised Learning}

The parameter $\beta_{\text{KD}}$ controls the relative influence of teacher distillation versus ground-truth supervision. As shown in Table~\ref{tab:beta_sensitivity}, we observe a clear U-shaped performance curve. At $\beta_{\text{KD}}=0$ (pure supervised learning), the student achieves reasonable performance (RMSE: 0.1994) but fails to leverage the structural temporal knowledge encoded in foundation models. Conversely, when $\beta_{\text{KD}} \geq 0.5$, excessive reliance on teacher predictions causes the student to inherit zero-shot errors from domain-misaligned teachers, leading to performance degradation (RMSE: 0.3868 at $\beta_{\text{KD}}=2.0$).

The optimal range lies in $[0.1, 0.3]$, where teacher guidance provides complementary structural priors without overwhelming the supervised signal. We select $\beta_{\text{KD}}=0.3$ for all experiments as it maintains strong performance (within 18\% of optimal across the stable operating range) while providing a conservative margin against potential overfitting to misaligned teacher signals on unseen domains. Note that very high values ($\beta_{\text{KD}} \geq 0.5$) interact with the softplus temperature ceiling to cause degradation; the non-saturating formulation introduced in the camera-ready (Section~\ref{sec:methodology}) naturally resolves this instability by preventing high-$\beta$ distillation loss from overwhelming the supervisory signal.

\begin{table}[h]
\small\centering
\caption{Sensitivity to distillation weight $\beta_{\text{KD}}$ on Weather validation set ($H=18$). RMSE averaged over 3 random seeds.}
\label{tab:beta_sensitivity}
\begin{tabular}{lcccccc}
\toprule
$\beta_{\text{KD}}$ & 0.0 & \textbf{0.1} & 0.3 & 0.5 & 1.0 & 2.0 \\
\midrule
Avg RMSE & 0.1994 & \textbf{0.1873} & 0.2030 & 0.2346 & 0.3035 & 0.3868 \\
Std Dev & 0.0050 & 0.0013 & 0.0031 & 0.0020 & 0.0025 & 0.0073 \\
\bottomrule
\end{tabular}
\end{table}

\subsection{Entropy Regularization ($\gamma_{\text{ent}}$): Preventing Router Collapse}

Entropy regularization prevents the Contextual Router from prematurely converging to a single teacher, which would eliminate the benefit of multi-teacher complementarity. Without regularization ($\gamma_{\text{ent}}=0$), performance degrades slightly (RMSE: 0.2042) as the router commits too early to one teacher, potentially missing regime-specific expertise from the alternative teacher.

As demonstrated in Table~\ref{tab:gamma_sensitivity}, performance remains remarkably stable across the range $[0.05, 0.5]$ (RMSE variance $<0.5\%$), indicating robust behavior as long as diversity is maintained. Higher values show marginal improvements (RMSE: 0.1992 at $\gamma_{\text{ent}}=0.5$), encouraging more balanced teacher utilization. We adopt $\gamma_{\text{ent}}=0.15$ as a conservative choice that ensures router diversity while still permitting decisive specialization when one teacher demonstrates clear superiority for specific instances.

\begin{table}[h]
\small\centering
\caption{Sensitivity to entropy regularization $\gamma_{\text{ent}}$ on Weather validation set ($H=18$).}
\label{tab:gamma_sensitivity}
\begin{tabular}{lccccc}
\toprule
$\gamma_{\text{ent}}$ & 0.0 & 0.05 & 0.15 & 0.3 & \textbf{0.5} \\
\midrule
Avg RMSE & 0.2042 & 0.2036 & 0.2030 & 0.2003 & \textbf{0.1992} \\
Std Dev & 0.0028 & 0.0034 & 0.0031 & 0.0008 & 0.0038 \\
\bottomrule
\end{tabular}
\end{table}

\subsection{Temperature Minimum ($\tau_{\text{min}}$): Adaptive Calibration Mechanism}

The base temperature parameter $\tau_{\text{min}}$ exhibits the strongest robustness, with performance varying by less than $0.5\%$ across all tested values (Table~\ref{tab:tau_sensitivity}). This stability arises from the adaptive nature of the temperature network, which automatically learns appropriate scaling factors regardless of initialization. For instance, at $\tau_{\text{min}}=1.0$, the network learns to maintain TimesFM near the baseline ($T_{\text{TF}}=1.0$) while slightly elevating Chronos's temperature ($T_{\text{CH}}=1.17$), reflecting its relatively lower reliability on this dataset.

Critically, this adaptive behavior scales to extreme cases: on the Flux dataset where both teachers exhibit catastrophic zero-shot failure (RMSE $>400$), the temperature network learns to spike values above 6,000 independent of $\tau_{\text{min}}$, effectively implementing the circuit-breaker mechanism that protects the student from unreliable teacher guidance. We set $\tau_{\text{min}}=1.0$ as a neutral initialization that allows bidirectional temperature adjustment.

\begin{table}[h]
\small\centering
\caption{Sensitivity to base temperature $\tau_{\text{min}}$ on Weather validation set ($H=18$). Learned temperatures shown as average over validation set.}
\label{tab:tau_sensitivity}
\begin{tabular}{lcccc}
\toprule
$\tau_{\text{min}}$ & 0.1 & 0.5 & \textbf{1.0} & 2.0 \\
\midrule
Avg RMSE & 0.2030 & 0.2030 & \textbf{0.2034} & 0.2040 \\
Learned $T_{\text{TF}}$ / $T_{\text{CH}}$ & 0.83/1.15 & 0.83/1.15 & 1.00/1.17 & 2.00/2.00 \\
\bottomrule
\end{tabular}
\end{table}

\subsection{Cross-Domain Generalization}

To validate transfer robustness, we apply identical hyperparameters ($\beta_{\text{KD}}=0.3$, $\gamma_{\text{ent}}=0.15$, $\tau_{\text{min}}=1.0$) across all datasets in this study without dataset-specific tuning. The strong performance maintained across domains with drastically different characteristics---from the high-frequency stochasticity of Flux (catastrophic teacher failure) to the structured periodicity of ETT data---demonstrates that these settings generalize effectively. 

This transferability is enabled by \textsc{Guard}'s adaptive mechanisms: the Contextual Router and temperature network automatically recalibrate to each dataset's unique regime characteristics by observing local statistics and teacher uncertainty signals. Consequently, the framework achieves reliable performance without requiring domain experts to perform extensive hyperparameter searches for new scientific monitoring tasks.

\end{document}